\documentclass[10pt,twocolumn,letterpaper]{article}

\usepackage{iccv}
\usepackage{times}
\usepackage{epsfig}
\usepackage{graphicx}
\usepackage{multirow}
\usepackage{amsmath}
\usepackage{amssymb}

\usepackage[pagebackref=true,breaklinks=true,letterpaper=true,colorlinks,bookmarks=false]{hyperref}

\iccvfinalcopy 

\def\iccvPaperID{28} 

\ificcvfinal\fi

\begin{document}

\title{Artist-Guided Semiautomatic Animation Colorization}
\author{Harrish Thasarathan ~and~ Mehran Ebrahimi\\
University of Ontario Institute of Technology\\
Oshawa, Ontario, Canada\\
{\tt\small {harrish.thasarathan@uoit.net  ~~~  mehran.ebrahimi}@uoit.net}\\
\small{\url{http://www.ImagingLab.ca}}
}

\maketitle
\ificcvfinal\thispagestyle{empty}\fi

\begin{abstract}
	There is a delicate balance between automating repetitive work in creative domains while staying true to an artist's vision. The animation industry regularly outsources large animation workloads to foreign countries where labor is inexpensive and long hours are common. Automating part of this process can be incredibly useful for reducing costs and creating manageable workloads for major animation studios and outsourced artists. We present a method for automating line art colorization by keeping artists in the loop to successfully reduce this workload while staying true to an artist's vision. By incorporating color hints and temporal information to an adversarial image-to-image framework, we show that it is possible to meet the balance between automation and authenticity through artist's input to generate colored frames with temporal consistency. 
	
\end{abstract}

\section{Introduction}

From multi-generational children's cartoons to modern adult classics, 2D animation continues to have a large impact on global television viewership. This is especially true with the recent growth of Japanese animation (anime). While traditional 2D animation was painstakingly hand drawn and painted frame by frame on thousands of pieces of parchment, its renaissance in the modern era has seen the field perfectly blend creativity with new technologies to expedite workflows. Even in its modern form, the frame by frame animation process remains a tedious endeavour. Developing these shows is a multistage process that requires multiple teams of artists simultaneously working on multiple tasks to meet production timelines. Most animation studios outsource large portions of their workloads to South Korea. When a single episode requires drawing and coloring between three to ten thousand frames, making corrections and alterations with overseas communication results in an 8-12 month production cycle per episode \cite{Anime:Fut}. The drawing of key frames that define major character movements are performed by lead artists while in-between frames that fill in these motions are animated by inexperienced artists. The colorization of line art sketches from key frames and in-between frames is considered to be tedious, repetitive, and low pay work. Thus finding an automatic pipeline to consistently produce thousands of colored frames from line art frames can be incredibly valuable to animation studios globally to expedite workflows.

\begin{figure}
    \centering
    \includegraphics[width=0.45\textwidth]{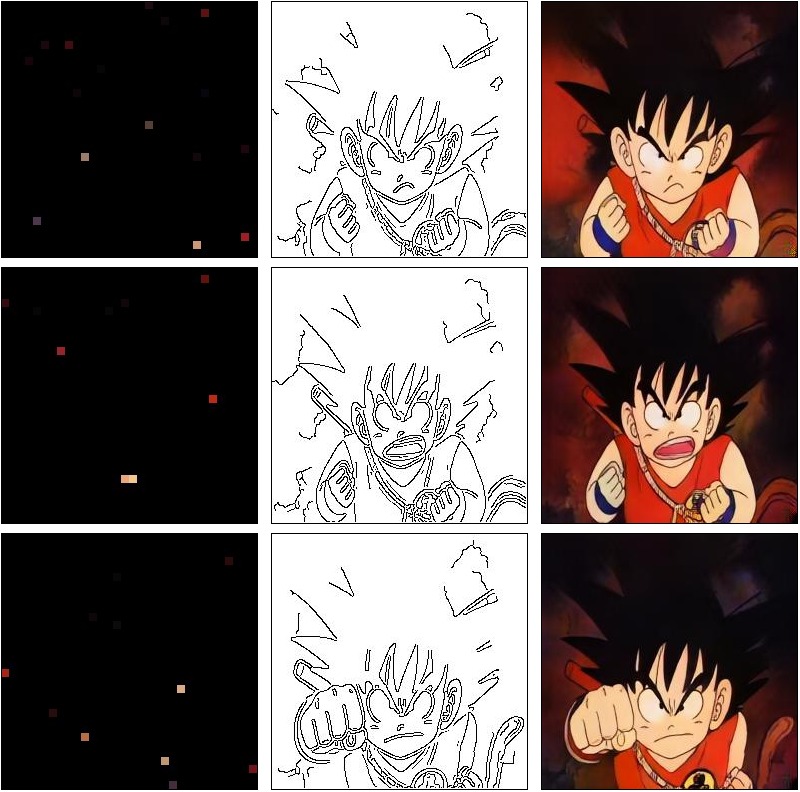}
    \caption{(Left) Color hint. (Center) Computed synthetic line art. (Right) Generated colorized temporally coherent frame from line art. }
    \label{fig:my_label}
\end{figure}


The image to image translation method presented by Isola \etal \cite{Pix2pix} uses Generative Adversarial Networks (GANs) conditioned on some input to learn a mapping from input to target image by minimizing a loss function. The mapping learned from the proposed technique works with some sparse inputs and has been used in applications such as converting segmentation labels to street view images. We aim to extend this model to work with synthetic line arts. Our model incorporates temporal information by encouraging consistency between colorized frames produced by the generator through discriminating against frame sequences. 

\section{Related Works}
Image to image translation using conditional GANs \cite{goodfellow2014generative,odena2018generator} is especially effective in comparison to CNN-based models for colorization tasks \cite{Nazeri:CGAN}. This model successfully maps a high dimensional input to a high dimensional output using a U-Net \cite{U-Net} based generator and patch-based discriminator \cite{Pix2pix}. The closer the input image is to the target, the better the learned mapping is. As a result, this technique is particularly suited to colorization tasks. User-guided colorization methods \cite{userhints} have also shown to be highly effective when the true color of an image is ambiguous and more accurate colors are required.

The neural algorithm for artistic style presented by Gatys \etal \cite{Style:Gatys} provides a method for the creation of artistic imagery. This is relevant as it demonstrates a way to learn representations of both content and style between two images using the pretrained VGG network \cite{simonyan2014very} and transfer that learned representation to a target image. The ability to learn and differentiate style and content using a pretrained network can supplement training for our purposes. 

There are very few existing methods for line art colorization. These methods are made to be highly generalized for use by independent artists and also incorporate scribble hints \cite{Anime:Paintschainer}. As a result, for production settings where sequential frames are related to each other, colorizing these frames independently creates a flicker effect from the inconsistencies between generated frames when made into a video. Additionally, providing detailed hints for every frame can be tedious. For our use case, a method that learns representations of recurring places and characters while keeping hints at a minimum is the most efficient in production.

\begin{figure*}
	\centering
	\includegraphics[height=.38\textheight]{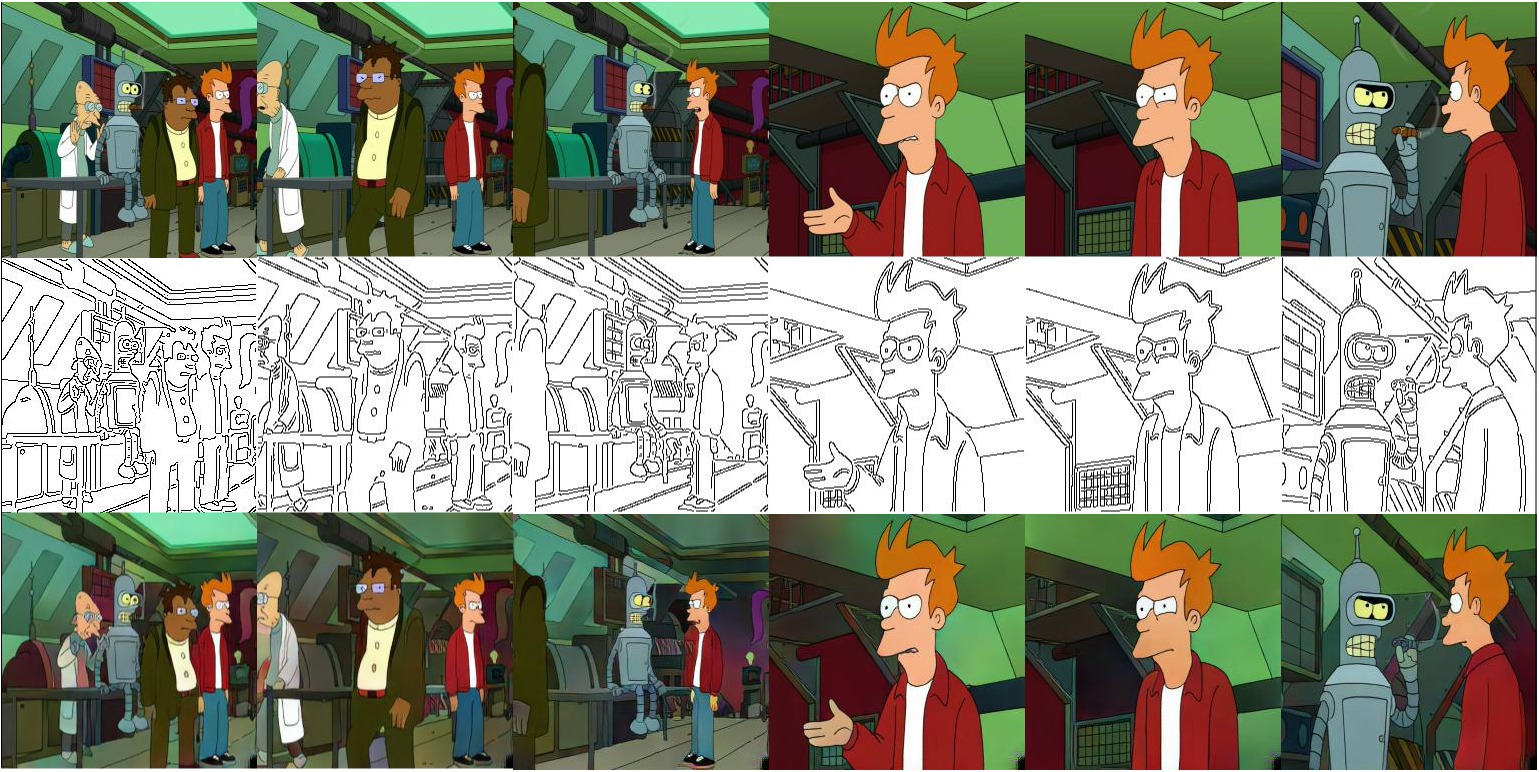}
	\caption{(Top) Ground truth frame sequence. (Middle) Line art frame sequence using Canny edge detection \cite{canny1986computational}. (Bottom) Colorized line art output from our proposed color hint and temporally coherent colorization model. The generated frame sequence has the same minor inconsistencies across frames showing that the temporal consistency method works as intended. The colorized output is accurate across different backgrounds.}
	\label{seq3}
\end{figure*}

\section{Model}
Our method attempts to colorize line art frames for large-scale production settings by taking into consideration temporal information to create consistency between frames. We also aim to keep artists in the loop by incorporating color hints. This allows for greater control by animators and accounts for variations that may be required during realistic production scenarios $i.e.$ new characters, new locations, existing characters wearing new clothes, etc. 

Our generator is comprised of $2$ downsampling layers followed by $8$ residual blocks \cite{residualblocks} and $2$ upsampling layers back to the input size. Instance normalization \cite{Norm:Instance} is used in place of batch normalization because we use smaller batches during training. Our discriminator architecture uses a $70\times70$ patch GAN as in Isola \etal \cite{Pix2pix} who map generator predictions to a scale of $N\times{N}$ outputs and classifies each patch as real or fake. Our discriminator also takes advantage of spectral normalization to stabilize training as shown in \cite{Norm:Spectral}.    

In Chan \etal \cite{Dance:Effros} temporal information is introduced in a cGAN architecture by adding extra conditions on their generator and discriminator. The current input and previously generated image provide conditions for their generator. The discriminator then differentiates between real and fake frame sequences instead of individual images. We take inspiration from this method and incorporate temporal information in our network with added conditions to our generator and discriminator.

\subsection{Loss Objective}
Let $G$ and $D$ represent the generator and discriminator of our colorization network. Our generator takes a greyscale or line art image $\mathbf{I}_{line}$ as input conditioned on the previous generated color frame prediction $\mathbf{F}_{t-1}$ and a color hint map $\mathbf{I}_{hint}$ to return a color prediction $\mathbf{F}_{t}$ temporally consistent to the previously colored frame prediction. 
\begin{equation}
    \mathbf{F}_{t} = G(\mathbf{I}_{line}, \mathbf{I}_{hint},\mathbf{F}_{t-1}).
\end{equation}
A joint loss is used to train the network that takes advantage of both conditional GAN and neural style algorithm that consists of an adversarial loss, style loss, content loss, and $l_{1}$ loss. The adversarial loss is defined in equation (\ref{eq:g_adv}) where the generator is encouraged to learn real sequences of colored frames because the discriminator differentiates sequences rather than individual images.
\begin{multline}
		\mathcal{L}_{adv} =\mathbb{E}_{(\mathbf{I}_{line},\mathbf{F}_{gt})} \left[ \log D(\mathbf{I}^{(t-1)}_{line},\mathbf{F}_{gt}^{(t-1)}, \mathbf{I}^{(t)}_{line}, \mathbf{F}_{gt}) \right] \\
		+ \mathbb{E}_{(\mathbf{I}_{line})} \log \left[ 1 - D(\mathbf{I}^{(t-1)}_{line},\mathbf{F}_{t-1}, \mathbf{I}^{(t)}_{line}, \mathbf{F}_{t}) \right].
	\label{eq:g_adv}
\end{multline}
We incorporate content and style loss described in \cite{Style:Gatys,Style:Johnson} to further supplement the training of our colorization network. Content loss encourages perceptual similarity while style loss encourages texture similarities between predicted and ground truth color frames. Perceptual similarity is accomplished by minimizing the Manhattan distance between feature maps generated by intermediate layers of VGG-19. Content loss is defined by equation (\ref{eq:l_content}) where $\phi_{i}$ represents the activation map at a given layer $i$ of VGG-19, $\mathbf{F}_{gt}$ is the current ground truth color frame, and $\mathbf{F}_{t}$ is the generated frame. $N_i$ represents the number of elements in the $i^{th}$ activation layer of VGG-19. For our work, we use activation maps from layers $\tt{relu1\_1, ~relu2\_1, ~relu3\_1, ~relu4\_1}$ and $\tt{relu5\_1}$ 

\begin{equation}
    \mathcal{L}_{cont} = \mathbb{E}_i \left[\frac{1}{N_i} \left\lVert \phi_i (\mathbf{F}_{gt}) - \phi_i (\mathbf{F}_{t}) \right \rVert_1 \right].
\label{eq:l_content}
\end{equation}
Style loss is calculated in a similar manner but rather than calculating the Manhattan distance between feature maps, we calculate the distance between the Gram matrices of the feature maps

\begin{equation}
   \mathcal{L}_{style} = \mathbb{E}_j \left[ \lVert G_j^{\phi} ({\mathbf{F}}_{t}) - G_j^{\phi} ({\mathbf{F}}_{gt}) \rVert_1 \right].
\label{eq:l_style}
\end{equation}

The Gram matrix of feature map $\phi$ is defined by $G^{\phi}_{j}$ in equation (\ref{eq:l_style}) which distributes spatial information containing non-localized information such as texture, shape, and style. We also add an $l_{1}$ term to our overall loss objective to preserve structure and encourage the generator to produce results similar to the ground truth. We use adversarial loss with $l_{1}$ to produce sharper generated outputs. The resulting final loss objective is the following 
\begin{equation}
    \mathcal{L} = \lambda_{adv}\mathcal{L}_{adv} + \lambda_{cont}\mathcal{L}_{cont}+ \lambda_{style}\mathcal{L}_{style}+ \lambda{l_{1}}\mathcal{L}_{l_{1}}.
\end{equation}

 For our experiments $\lambda_{adv}$ = $\lambda_{content}$ = 1, $\lambda_{style}$ = 1000 and $\lambda_{l1}$ = 10. We do not incorporate content loss for greyscale experiments since the content of a greyscale and colored image are the same.
 
\begin{table*}
\centering
\begin{tabular}{r|c|c|c|c|c|c|c|c}
\hline
\multirow{3}{*}{Statistic} & \multicolumn{8}{c}{Model}                               \\ 
                           & \multicolumn{2}{c}{DragonBall (Baseline)} & \multicolumn{2}{c}{DragonBall (Ours)} &\multicolumn{2}{c}{Futurama (Baseline)} &\multicolumn{2}{c}{Futurama (Ours)}\\ \cline{2-9}
                           & Greyscale      & Line art     & Greyscale    & Line art   &Greyscale &Lineart & Greyscale & Line art\\ \hline\hline 
FID                        & 9.29          &  19.1            &2.48        & 8.84     &6.14 &9.50  & 3.51 & 9.46  \\ \hline
SSIM                       & 0.78            & 0.57       & 0.90    & 0.73           &0.79 & 0.65 & 0.87 & 0.68\\ \hline
PSNR                       & 17.4              & 16.04           & 24.9  & 19.57     &20.6 & 18.1  &24.8 &19.4\\ \hline
\end{tabular}
\caption{Quantitative results Structure Similarity Index (SSIM), Peak Signal to Noise Ratio (PSNR), and Fr\'echet Inception Distance (FID) between results of a  baseline without hints conditioned with $F_{gt}^{(t-1)}$ and our proposed model on Futurama and Dragonball datasets. }\label{tab1}
\end{table*}

\section{Experiments}
\subsection{Dataset}
In order to show viability for large-scale productions, both for Japanese animation and traditional Western animation, we create two datasets by extracting frames from DragonBall and Futurama. We extract frames from seasons 1 and 2 from both shows using OpenCV to create datasets of $100,000$ images for each. To mimic line art we use Canny edge detection \cite{canny1986computational}.

\subsection{Training}
During training, the first $\mathbf{F}_{t-1}$ is a blank image. In our application scenario, animators can choose patches of color and drag them onto the line art image to have more control over the colorization output of the network. To mimic this during training, we convert the ground truth image into a set of $4\times4$ patches of the averaged color in the region. One percent of these patches from random locations are shown to the network against a black background as the color hint.

\section{Evaluation}
Results in Table \ref{tab1} obtained from an existing baseline model trained on the Dragonball dataset with the previous ground truth frame as a condition in \cite{thasarathan2019tcvc}, still suffers from the flicker effect due to inconsistencies in color between subsequent frames. In addition, unfamiliar backgrounds and characters suffered the most as the model colored them differently for each frame. On the other hand, when incorporating user-hints and temporal information, we can both achieve higher level of color consistency between related frames as well as color new objects unseen by the network previously with better accuracy. This is also reflected in our quantitative results where our model performs better than the baseline across each metric. The score similarity between our model trained on the Futurama and Dragonball datasets highlight the efficacy of our approach for both Western and Japanese animation styles. Our method makes a more significant difference when colorizing Japanese animation styles where color and shading is more complex versus flat colors and shading used in Western animation.

\section{Discussion and Future Work}

It is evident from our results that the proposed method is a viable solution to heavy colorization workloads. By incorporating color hints, large-scale production animations can reduce the amount of time that is allocated for colorization, while keeping artists in control over an automation pipeline. This allows for the artist's desired color results when new settings and characters are introduced to a series. By incorporating a temporal condition to our adversarial framework, we encourage the generator to produce temporally coherent frames by classifying generated frame sequences instead of images with our discriminator. We notice that simply conditioning the generator on the ground truth image from the previous time step so that the network learns to copy over colors from the generated image at test time is not as effective. Future work can include a more extensive model to effectively replicate line art during training. 


\section*{Acknowledgments} This research was supported by the NSERC Discovery Grant for M.E.~ H.T. is supported by an institutional Undergraduate Student Research Fellowship. We gratefully acknowledge the support of NVIDIA Corporation for the donation of GPUs used in this research.
{\small
\bibliographystyle{ieee}
\bibliography{egbib}
}

\end{document}